\documentclass[sigconf]{acmart}

\usepackage{booktabs} 
\usepackage{subcaption}
\usepackage{microtype}
\usepackage{cleveref}
\usepackage{float}
\usepackage{hyperref}

\AtBeginDocument{%
  \providecommand\BibTeX{{%
    \normalfont B\kern-.05em{\scshape i\kern-.025em b}\kern-.08em \TeX}}}

\copyrightyear{2026}
\acmYear{2026}
\setcopyright{cc}
\setcctype{by}
\acmConference[RecSys '26]{20th ACM Conference on Recommender Systems}{September 27-October 02, 2026}{Minneapolis, MN, USA}
\acmBooktitle{20th ACM Conference on Recommender Systems (RecSys '26), September 27-October 02, 2026, Minneapolis, MN, USA}
\acmDOI{10.1145/3773078.3841273}
\acmISBN{979-8-4007-2284-4/2026/09}

\begin{document}

\title{AutoRecLab: Describe the Experiment, Get the Code!}

\author{Moritz Baumgart}
\affiliation{%
  \institution{University of Siegen}
  \city{Siegen}
  \country{Germany}}
\email{moritz.baumgart@uni-siegen.de}
\orcid{0009-0007-1322-1450}

\author{Philipp Meister}
\affiliation{%
  \institution{University of Siegen}
  \city{Siegen}
  \country{Germany}}
\email{philipp.meister@uni-siegen.de}
\orcid{0009-0008-6814-9668}

\author{Justus Krell}
\affiliation{%
  \institution{University of Siegen}
  \city{Siegen}
  \country{Germany}}
\email{justus.krell@student.uni-siegen.de}
\orcid{0009-0004-2234-3749}

\author{Michael Schmidt}
\affiliation{%
  \institution{University of Siegen}
  \city{Siegen}
  \country{Germany}}
\email{michael3.schmidt@student.uni-siegen.de}
\orcid{0009-0000-4862-0450}

\author{Bela Gipp}
\affiliation{%
  \institution{University of Göttingen}
  \city{Göttingen}
  \country{Germany}}
\email{bela.gipp@uni-goettingen.de}
\orcid{0000-0001-6522-3019}

\author{Joeran Beel}
\affiliation{%
  \institution{University of Siegen}
  \city{Siegen}
  \country{Germany}}
\email{joeran.beel@uni-siegen.de}
\orcid{0000-0002-4537-5573}

\begin{abstract}
Empirical evaluation is central to recommender-systems (RecSys) research, but turning experimental designs into executable code remains a manual and error-prone task. We present AutoRecLab, a Python-based autonomous RecSys lab that automates RecSys experiments from natural-language prompts. Given a research idea, AutoRecLab derives explicit experiment requirements, builds and validates a prototype, and iteratively expands it into the requested full experiment. The workflow combines retrieval-augmented generation (RAG) for documentation lookup, static type verification, and execution-steered tree search. In our demonstration, AutoRecLab autonomously implements an explicit-to-implicit feedback conversion study. In a baseline comparison across six algorithms and three datasets, 8 of 9 runs succeed at an average cost of approximately \$1 per run with GPT-5.4-mini.
\end{abstract}

\ccsdesc[500]{Information systems~Recommender systems}
\ccsdesc[300]{Computing methodologies~Natural language generation}

\keywords{Recommender Systems, Autonomous Agents, Code Generation, RAG, LLM}

\maketitle

\section{Introduction}
Autonomous science agents based on Large Language Models (LLMs) increasingly automate science and engineering tasks specified through natural-language prompts~\cite{Lu2024, Yamada2025}. Recommender-systems (RecSys) research is a suitable domain for this type of automation because setting up and implementing experiments requires considerable manual work. Researchers often need to write custom preprocessing code, configure evaluation loops, and adapt implementations to libraries such as LensKit~\cite{ekstrand2020lenskit}, RecBole~\cite{zhao2021recbole}, or meta-frameworks such as OmniRec~\cite{wegmeth2026omnirec}.

Learning these libraries can help researchers understand experimental choices, but it also adds setup work for newcomers. Experienced researchers likewise spend considerable time implementing and debugging experiments, and manual implementations remain susceptible to evaluation and reproducibility errors across libraries~\cite{FerrariDacrema2019, Beel2024Best}. Systems that turn high-level research ideas into explicit experiment requirements can reduce routine implementation work and check generated code against those requirements. RecSys adds a methodological reason for such support: choices such as data filtering, splitting, candidate construction, and random seeds can materially affect experimental outcomes and their interpretation~\cite{Beel2016,wegmeth2023effect,Beel2024Best}. A general code-generation agent can therefore return executable software without reliably preserving the intended evaluation protocol. AutoRecLab makes the translation from a research request to an executable experiment explicit and inspectable and combines it with RecSys-specific software and validation steps.

Standard LLMs can assist researchers, but they frequently hallucinate API calls for domain-specific RecSys libraries, and existing general science agents do not include RecSys-specific configuration. We examined several LLM-based science agents, including Sakana's AI Scientist~\cite{Lu2024,Yamada2025}, Agent Laboratory~\cite{Schmidgall2025}, AI-Researcher~\cite{Tang2025}, and Zochi~\cite{Intology2025Zochi}, to assess their use for recommender-systems research. The agents we could test required additional context because they focus primarily on machine-learning tasks. Our detailed evaluation of AI Scientist likewise found that its performance on RecSys tasks fell short of expectations~\cite{Beel2025b}.

Our work on AutoRecLab builds on our earlier independent evaluation of Sakana's AI Scientist in recommender-systems research~\cite{Beel2025b}. That study found substantial limitations in literature review, experiment execution, and methodological correctness: five of twelve proposed experiments failed because of coding errors, while several executable experiments still produced flawed or misleading results. These findings motivated the development of research agents that incorporate RecSys-specific software, experimental knowledge, and validation procedures.

Our group publicly introduced the AutoRecLab concept in October 2025~\cite{beel2025autorecsys}, and subsequent work further developed the agenda for automated RecSys research~\cite{beel2026autoreclabs}. To the best of our knowledge, AutoRecLab is the first publicly documented open-source research agent developed specifically for recommender-systems experimentation.\footnote{\url{https://github.com/ISG-Siegen/AutoRecLab}} It takes natural-language RecSys research tasks through requirement derivation, code generation, execution, evaluation, and iterative refinement.

The implementation described in this paper is an early proof of concept of that vision. Starting from a single natural-language prompt, AutoRecLab derives explicit experiment requirements, builds and validates a small prototype, and then refines it into the requested full experiment. This staged process reduces the cost of detecting implementation errors before the complete experiment is executed and checks each generated implementation against the experiment-specific requirements.

\section{Related Work}
AI Scientist systems now automate large parts of computational research workflows. Sakana's AI Scientist generates ideas, implements and runs experiments, and drafts manuscripts; Agent Laboratory starts from a human-provided idea and automates literature review, experimentation, and report writing~\cite{Lu2024,Yamada2025,Schmidgall2025}. AI-Researcher targets end-to-end scientific innovation, Data-to-Paper turns data and analyses into human-verifiable papers, and CodeScientist links idea generation with code-based experimentation~\cite{Tang2025,Ifargan2025,allenai2025codescientist}. These systems provide evidence that agentic research workflows are feasible, although their development and evaluation have focused mainly on general or machine-learning-oriented research instead of the methodological and software conventions of recommender-systems experiments.

A second line of work concentrates on autonomous machine-learning engineering and experimentation. MLAgentBench tests language-model agents on iterative ML experiments, MLE-bench evaluates agents across 75 Kaggle competitions, and AIDE treats ML engineering as tree search over executable code~\cite{Huang2023MLAgentBench,Chan2024MLEBench,Jiang2025AIDE}. These systems are relevant to AutoRecLab because they use execution feedback, repeated refinement, and search over candidate implementations to automate experimentation. Their benchmarks primarily measure successful ML engineering or performance improvement and do not cover RecSys-specific choices such as interaction preprocessing, candidate construction, ranking evaluation, or compatibility across recommendation libraries.

Recommender-systems research has its own history of experiment infrastructure and partial automation. LensKit, RecBole, RecPack, and Elliot standardize parts of data processing, recommendation, and evaluation, while OmniRec supplies a common layer across several libraries~\cite{ekstrand2020lenskit,zhao2021recbole,michiels2022recpack,anelli2021elliot,wegmeth2026omnirec}. Auto-Surprise and LensKit-Auto also automate algorithm selection and hyperparameter optimization inside predefined RecSys pipelines~\cite{Anand2020AutoSurprise,Vente2023LensKitAuto}. AutoRecLab adds a research-automation layer above these tools: it translates a research request into explicit requirements, builds and runs the corresponding experiment, and iteratively checks and refines the implementation. Our earlier position work described this transition from AutoRecSys toward autonomous RecSys research~\cite{beel2025autorecsys,beel2026autoreclabs}.

\section{AutoRecLab}
AutoRecLab is an open-source Python command-line tool that converts a natural-language research prompt into valid, executable code. Its workflow has three phases: requirements engineering, prototyping, and refinement from the prototype to the complete experiment. AutoRecLab acts as a research-automation layer above the recommender algorithms and experimentation libraries. The prompt supplies the research specification, which AutoRecLab converts into machine-checkable requirements. An LLM then generates candidate implementations, retrieved documentation grounds API use, and static checks and execution feedback guide code improvement. OmniRec provides the execution layer that connects the generated experiment to datasets and recommendation libraries. Separating a small prototype from the complete experiment lets AutoRecLab establish an executable implementation before expanding it to the full requested study. During these phases, AutoRecLab checkpoints intermediate states and stores Python code, generated plots, execution logs, and other artifacts in a dedicated workspace. For experiment execution, it uses the OmniRec~\cite{wegmeth2026omnirec} meta-framework, which standardizes data loading and training across more than 230 datasets and several RecSys Python libraries: RecPack~\cite{michiels2022recpack}, RecBole~\cite{zhao2021recbole}, LensKit~\cite{ekstrand2020lenskit}, and Elliot~\cite{anelli2021elliot}.

\subsection{Demonstration}
AutoRecLab is distributed as a local tool, so no live-system link is provided. The source code and development history are available in the public GitHub repository\footnote{\url{https://github.com/ISG-Siegen/AutoRecLab}.\label{github}}, from which users can run AutoRecLab on their own machines.

\begin{figure}[htbp]
    \centering
    \includegraphics[width=0.9\linewidth]{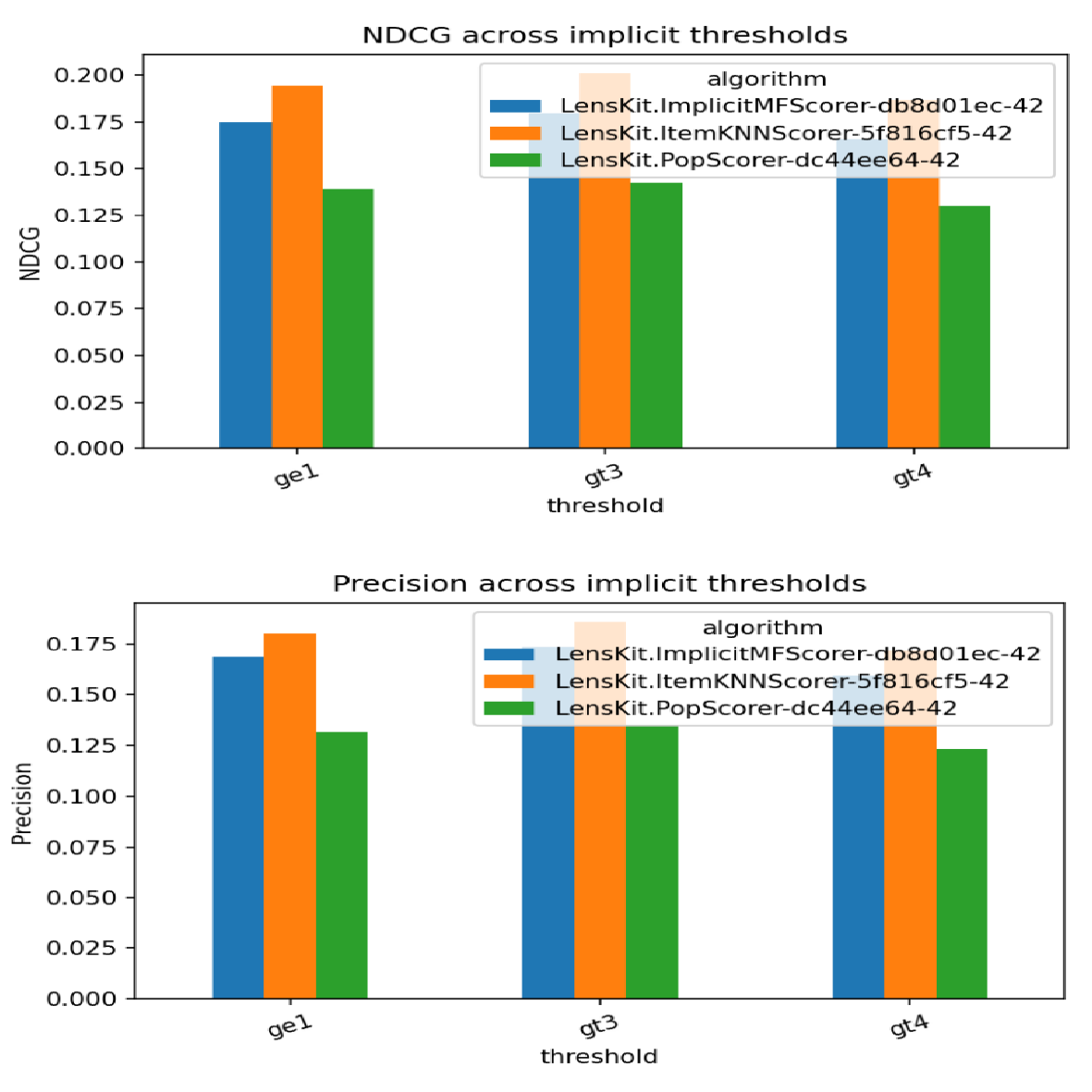}
    \caption{Plot generated by AutoRecLab for the explicit-to-implicit conversion experiment. The panels report NDCG@10 (top) and Precision@10 (bottom) on MovieLens 1M for different algorithms and rating thresholds: greater than or equal to (ge) 1, greater than 3, and greater than 4.}
    \Description{Two vertically stacked grouped bar charts, each with three bars per feedback-conversion condition.
    For NDCG@10, ItemKNN is highest when ratings greater than or equal to 1 are converted to implicit feedback, whereas ImplicitMF is highest for thresholds greater than 3 and greater than 4.
    For Precision@10, ItemKNN is highest for ratings greater than or equal to 1 and greater than 3, whereas ImplicitMF is highest for ratings greater than 4.
    The popularity baseline is lowest in every condition, and all three algorithms reach their lowest values when only ratings greater than 4 are converted.}
    \label{fig:generated_plot}
\end{figure}

We evaluated AutoRecLab in four representative empirical RecSys scenarios. The demonstration focuses on an explicit-to-implicit feedback conversion experiment with the MovieLens 1M dataset, initiated by the following prompt:
\begin{quote}
    \textit{
    Test the influence of [...] feedback conversion strategies on recommendation accuracy by comparing multiple binarization thresholds [...].
    Evaluate [...] on the MovieLens1M dataset.
    Report metrics [...], and compare ranking quality [...].}
\end{quote}
From this prompt, AutoRecLab derived 22 requirements covering data loading, conversion thresholds, train-test splitting, and evaluation. It produced 169 lines of executable Python code and the requested comparative plots (see \cref{fig:generated_plot}) for a total API cost of USD 0.76 with \texttt{GPT-5.4-mini}.

To assess reproducibility across standard algorithms, we asked AutoRecLab to establish a performance baseline with 6 algorithms for model comparison. Eight of the nine runs ($\approx 89\%$) generated bug-free code and plots, with a cost of about \$1 per run. \Cref{tab:autoreclab_results} reports the run-level statistics. Some runs have long total runtimes because execution of the generated code dominates the elapsed time.

\begin{table}[htbp]
    \centering
    \caption{Statistics for the AutoRecLab baseline experiment across nine runs (P: Prototype, F: Final Refinement)}
    \label{tab:autoreclab_results}
    \scriptsize 
    \setlength{\tabcolsep}{2.5pt} 
    \begin{tabular}{l *{9}{c}}
    \toprule
    & \multicolumn{9}{c}{\textbf{Run Number}} \\
    \textbf{Metric} & \textbf{1.1} & \textbf{1.2} & \textbf{1.3} & \textbf{2.1} & \textbf{2.2} & \textbf{2.3} & \textbf{3.1} & \textbf{3.2} & \textbf{3.3} \\
    \midrule
    \textbf{Dataset}      & MU      & MU       & MU      & ML1M    & ML1M    & ML1M    & VI      & VI      & VI      \\
    \textbf{Algos}        & 6       & 6        & 6       & 6       & 6       & 6       & 6       & 6       & 6       \\
    \textbf{Cost (\$)}    & 1.09    & 1.08     & 1.10    & 0.95    & 0.98    & 0.91    & 1.04    & 1.01    & 0.90    \\
    \textbf{Runtime P}    & 4.37h  & 12.23h  & 40m     & 23m     & 58m     & 1.52h & 17.32h & 7m      & 19m     \\
    \textbf{Runtime F}    & 30.2h   & 33.65h  & 16.23h  & 32.33h & 32.08h  & 22.43h  & 5m      & 24.48h & 16.58h  \\
    \textbf{Best Score P} & 0.89    & 0.84     & 0.375   & 0.6     & 0.875   & 0.857   & 0.56    & 0.45    & 0.4     \\
    \textbf{Avg. LoC}     & 110     & 111      & 139     & 100     & 134     & 118     & 100     & 141     & 109     \\
    \textbf{Nodes P}      & 8       & 8        & 8       & 8       & 8       & 8       & 8       & 8       & 8       \\
    \textbf{Nodes F}      & 4       & 4        & 4       & 4       & 4       & 4       & 4       & 4       & 4       \\
    \textbf{Buggy P?}     & No      & No       & Yes     & No      & No      & No      & Yes     & Yes     & Yes     \\
    \textbf{Buggy F?}     & No      & No       & No      & No      & No      & No      & Yes     & No      & No      \\
    \bottomrule
    \end{tabular}
    
    \smallskip
    \scriptsize
    \raggedright
    \textbf{Datasets:} \textbf{ML1M}: MovieLens 1M, \textbf{MU}: Amazon2018MusicalInstruments, \textbf{VI}: Amazon2018VideoGames.
\end{table}

We also evaluated AutoRecLab on a dataset-filtering task that measured the performance effect of pruning users with few interactions. A final scenario examined how random seeds used for user splitting affect evaluation metrics. The resulting patterns reproduced qualitative trends from existing human-conducted research~\cite{wegmeth2023effect}, indicating that AutoRecLab can support empirical RecSys experimentation.

\subsection{System Architecture}
\begin{figure}[htbp]
    \centering
    \includegraphics[width=0.85\linewidth]{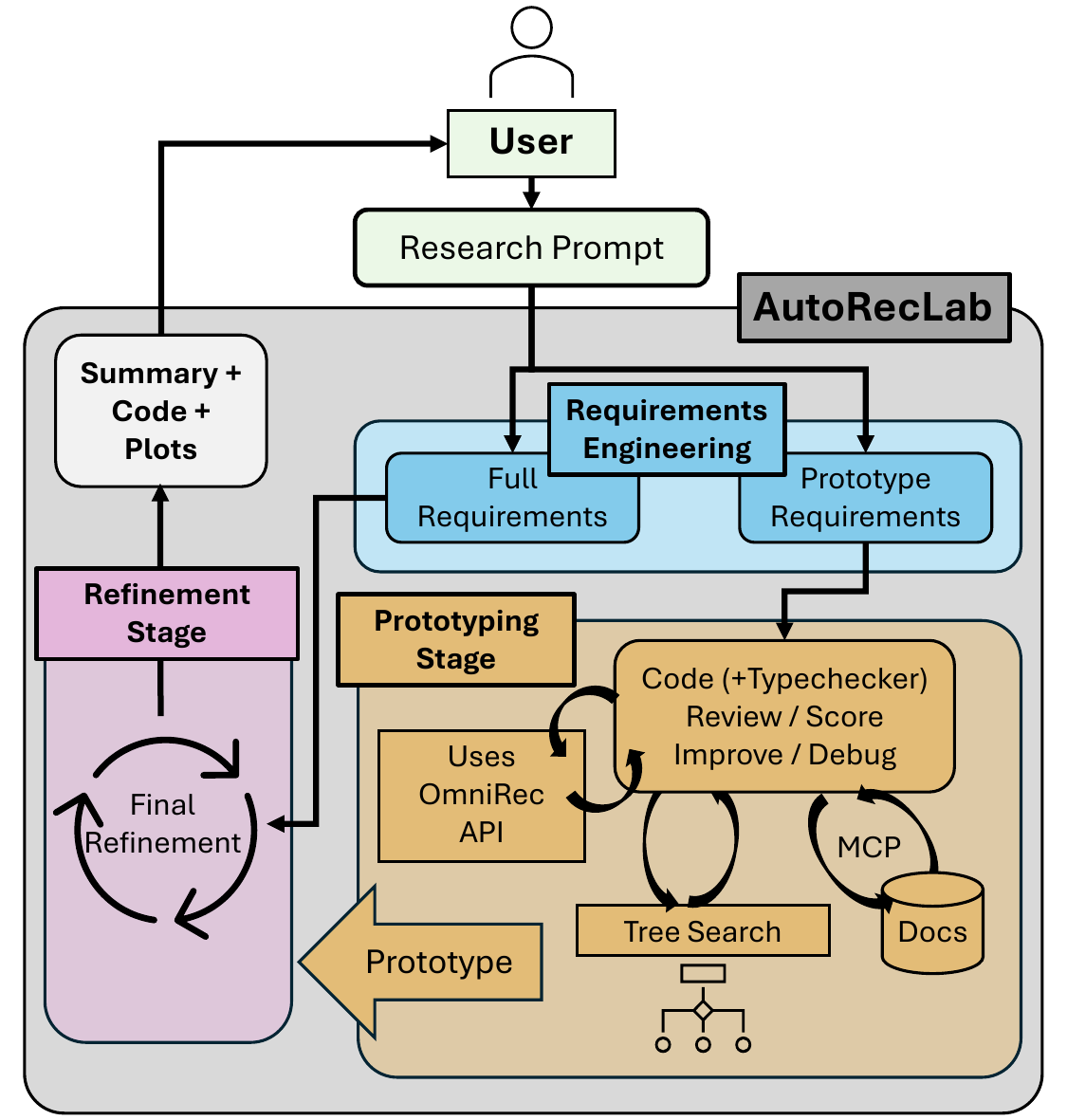}
    \caption{AutoRecLab workflow. Requirements are derived from the research task. During prototyping, tree search and MCP-based documentation retrieval produce a working prototype. The refinement loop expands this prototype to the full experiment and returns the experimental summary, code, and plots.}
    \Description{A block diagram of the AutoRecLab workflow.
    A user provides a Research Prompt, which enters Requirements Engineering and is separated into Full and Prototype Requirements.
    Prototype Requirements enter the Prototyping Stage, where code is iterated with the OmniRec API, Tree Search, and MCP-assisted documentation retrieval.
    The resulting Prototype enters the Refinement Stage and is combined with the Full Requirements in a Final Refinement loop that returns a summary, code, and plots to the user.}
    \label{fig:architecture}
\end{figure}

AutoRecLab implements and executes experiments through the three stages shown in \cref{fig:architecture}.

\subsubsection{Requirements Engineering}
AutoRecLab translates the user's natural-language prompt into a research plan and two requirement sets. \textbf{(1) Prototype Requirements} specify a small, fast-running experiment, usually limited to one dataset, one baseline algorithm, and one metric cutoff. \textbf{(2) Full Requirements} preserve the complete request, including every specified algorithm, dataset, metric, and visualization.

This separation lets AutoRecLab test the implementation on a small experiment before expanding it to the full setup.

\subsubsection{Prototyping Stage}
AutoRecLab currently executes experiments through OmniRec and is therefore limited to the libraries and datasets that OmniRec supports. Using AutoRecLab reduces manual setup; using OmniRec directly gives researchers more immediate control over configuration and implementation.

\textbf{Code Generation \& Verification}: AutoRecLab generates code and applies static checks for type mismatches. Detected errors start an automated correction loop.

\textbf{RAG Documentation Server (MCP)}: AutoRecLab reduces hallucinated API calls by retrieving indexed documentation and code for OmniRec, LensKit, and RecBole through the Model Context Protocol (MCP).

\textbf{Evaluation \& Tree Search}: AutoRecLab executes each candidate in an independent workspace. An LLM evaluates the generated code and console output and classifies, for every prototype requirement, whether the candidate fulfills it and whether the candidate is buggy or bug-free. The fraction of fulfilled requirements defines the node score $S \in [0,1]$. This score records requirement coverage and does not represent general confidence or guarantee correctness. Candidate implementations form a search tree. For node selection, AutoRecLab first chooses whether to sample from the buggy or bug-free node set. An $\epsilon$-greedy strategy then selects either the highest-scoring candidate in that set or a random alternative for improvement or debugging. The search stops when a candidate reaches $S=1$ or the configured iteration limit is reached.

\subsubsection{Refinement Stage}

AutoRecLab starts from the executable prototype and incrementally extends it until the full requirements are satisfied. Each revision is executed and evaluated before the next refinement. At the end of the process, AutoRecLab returns the Python code, generated plots, execution logs, and a Markdown summary for inspection and modification.

\section{Conclusion}
The evaluation covers a limited set of comparatively simple offline RecSys tasks. Across nine runs, AutoRecLab produced code classified as bug-free in eight cases, although some resulting analyses were not scientifically meaningful. The results suggest that the approach is technically feasible and indicate that requirement coverage and successful execution alone may not fully capture experiment quality.

AutoRecLab is an early proof of concept for the broader vision of autonomous RecSys research labs~\cite{beel2025autorecsys}. Its current capabilities depend on OmniRec, the underlying LLM, the coverage of indexed documentation, and a sequential search process that can produce long runtimes. For supported tasks, AutoRecLab can reduce implementation effort and produce inspectable artifacts; researchers remain responsible for experimental design, code inspection, result interpretation, and decisions about when user studies are required. Future evaluations can cover more complex RecSys tasks and different LLMs, assess generated code quality explicitly, index additional recommendation libraries, and parallelize the tree search. Further extensions could add support for literature search and manuscript preparation.

\bibliographystyle{ACM-Reference-Format}
\bibliography{references}

@inproceedings{anelli2021elliot,
  title={Elliot: A comprehensive and rigorous framework for reproducible recommender systems evaluation},
  author={Anelli, Vito Walter and Bellog{\'\i}n, Alejandro and Ferrara, Antonio and Malitesta, Daniele and Merra, Felice Antonio and Pomo, Claudio and Donini, Francesco Maria and Di Noia, Tommaso},
  booktitle={Proceedings of the 44th international ACM SIGIR conference on research and development in information retrieval},
  pages={2405--2414},
  year={2021}
}

@inproceedings{michiels2022recpack,
  title={Recpack: An (other) experimentation toolkit for top-n recommendation using implicit feedback data},
  author={Michiels, Lien and Verachtert, Robin and Goethals, Bart},
  booktitle={Proceedings of the 16th ACM Conference on Recommender Systems},
  pages={648--651},
  year={2022}
}

@inproceedings{zhao2021recbole,
  title={Recbole: Towards a unified, comprehensive and efficient framework for recommendation algorithms},
  author={Zhao, Wayne Xin and Mu, Shanlei and Hou, Yupeng and Lin, Zihan and Chen, Yushuo and Pan, Xingyu and Li, Kaiyuan and Lu, Yujie and Wang, Hui and Tian, Changxin and others},
  booktitle={proceedings of the 30th acm international conference on information \& knowledge management},
  pages={4653--4664},
  year={2021}
}

@inproceedings{ekstrand2020lenskit,
  title={Lenskit for python: Next-generation software for recommender systems experiments},
  author={Ekstrand, Michael D},
  booktitle={Proceedings of the 29th ACM international conference on information \& knowledge management},
  pages={2999--3006},
  year={2020}
}

@inproceedings{wegmeth2026omnirec,
  title={OmniRec: The All-In-One Solution for Reproducible and Interoperable Recommender Systems Experimentation},
  author={Wegmeth, Lukas and Baumgart, Moritz and Meister, Philipp and Gipp, Bela and Beel, Joeran},
  booktitle={European Conference on Information Retrieval},
  pages={129--135},
  year={2026},
  organization={Springer}
}

@Misc{Lu2024,
  author    = {Lu, Chris and Lu, Cong and Lange, Robert Tjarko and Foerster, Jakob and Clune, Jeff and Ha, David},
  title     = {The AI Scientist: Towards Fully Automated Open-Ended Scientific Discovery},
  year      = {2024},
  copyright = {Creative Commons Attribution 4.0 International},
  doi       = {10.48550/ARXIV.2408.06292},
  publisher = {arXiv},
}

@Misc{Yamada2025,
  author    = {Yamada, Yutaro and Lange, Robert Tjarko and Lu, Cong and Hu, Shengran and Lu, Chris and Foerster, Jakob and Clune, Jeff and Ha, David},
  title     = {The AI Scientist-v2: Workshop-Level Automated Scientific Discovery via Agentic Tree Search},
  year      = {2025},
  copyright = {Creative Commons Attribution 4.0 International},
  doi       = {10.48550/ARXIV.2504.08066},
  publisher = {arXiv},
}

@Misc{Schmidgall2025,
  author    = {Schmidgall, Samuel and Su, Yusheng and Wang, Ze and Sun, Ximeng and Wu, Jialian and Yu, Xiaodong and Liu, Jiang and Moor, Michael and Liu, Zicheng and Barsoum, Emad},
  title     = {Agent Laboratory: Using LLM Agents as Research Assistants},
  year      = {2025},
  copyright = {Creative Commons Attribution 4.0 International},
  doi       = {10.48550/ARXIV.2501.04227},
  publisher = {arXiv},
}

@Misc{Tang2025,
  author    = {Tang, Jiabin and Xia, Lianghao and Li, Zhonghang and Huang, Chao},
  title     = {AI-Researcher: Autonomous Scientific Innovation},
  year      = {2025},
  copyright = {Creative Commons Attribution 4.0 International},
  doi       = {10.48550/ARXIV.2505.18705},
  publisher = {arXiv},
}

@InProceedings{wegmeth2023effect,
  author    = {Wegmeth, Lukas and Vente, Tobias and Purucker, Lennart and Beel, Joeran},
  booktitle = {Perspectives@ RecSys},
  title     = {The Effect of Random Seeds for Data Splitting on Recommendation Accuracy.},
  year      = {2023},
}

@InProceedings{FerrariDacrema2019,
  author     = {Ferrari Dacrema, Maurizio and Cremonesi, Paolo and Jannach, Dietmar},
  booktitle  = {Proceedings of the 13th ACM Conference on Recommender Systems},
  title      = {Are we really making much progress? A worrying analysis of recent neural recommendation approaches},
  year       = {2019},
  month      = sep,
  pages      = {101--109},
  publisher  = {ACM},
  series     = {RecSys ’19},
  collection = {RecSys ’19},
  doi        = {10.1145/3298689.3347058},
}

@Article{Beel2016,
  author    = {Beel, Joeran and Breitinger, Corinna and Langer, Stefan and Lommatzsch, Andreas and Gipp, Bela},
  journal   = {User Modeling and User-Adapted Interaction},
  title     = {Towards reproducibility in recommender-systems research},
  year      = {2016},
  issn      = {1573-1391},
  month     = mar,
  number    = {1},
  pages     = {69--101},
  volume    = {26},
  doi       = {10.1007/s11257-016-9174-x},
  publisher = {Springer Science and Business Media LLC},
}

@inproceedings{Beel2024Best,
    title = {Best-Practices for Offline Evaluations of Recommender Systems},
    author = {Joeran Beel and Dietmar Jannach and Alan Said and Guy Shani and Tobias Vente and Lukas Wegmeth},
    editor = {Christine Bauer and Alan Said and Eva Zangerle},
    year = {2024},
    date = {2024-01-01},
    booktitle = {Report from Dagstuhl Seminar 24211 – Evaluation Perspectives of Recommender Systems: Driving Research and Education},
    pubstate = {published},
    tppubtype = {inproceedings}
}

@Article{Ifargan2025,
  author    = {Ifargan, Tal and Hafner, Lukas and Kern, Maor and Alcalay, Ori and Kishony, Roy},
  journal   = {NEJM AI},
  title     = {Autonomous LLM-Driven Research — from Data to Human-Verifiable Research Papers},
  year      = {2025},
  issn      = {2836-9386},
  month     = jan,
  number    = {1},
  volume    = {2},
  doi       = {10.1056/aioa2400555},
  publisher = {Massachusetts Medical Society},
}

@Misc{allenai2025codescientist,
  author       = {{Allen Institute for AI}},
  howpublished = {\url{https://github.com/allenai/codescientist}},
  note         = {Accessed: 2025-10-20},
  title        = {Code Scientist},
  year         = {2025},
}

@Article{Beel2025b,
  author    = {Beel, Joeran and Kan, Min-Yen and Baumgart, Moritz},
  journal   = {ACM SIGIR Forum},
  title     = {Evaluating Sakana’s AI Scientist: Bold Claims, Mixed Results, and a Promising Future?},
  year      = {2025},
  issn      = {0163-5840},
  month     = jun,
  number    = {1},
  pages     = {1--20},
  volume    = {59},
  doi       = {10.1145/3769733.3769747},
  publisher = {Association for Computing Machinery (ACM)},
}

@Misc{Intology2025Zochi,
  author       = {Zhou, Andy and Arel, Ron and Dunn, Soren and Khandekar, Nikhil},
  howpublished = {\url{https://www.intology.ai/blog/zochi-tech-report}},
  month        = {mar},
  note         = {Accessed: 2025-10-20},
  title        = {Zochi Technical Report},
  year         = {2025},
}

@article{beel2025autorecsys,
  title={From AutoRecSys to AutoRecLab: A Call to Build, Evaluate, and Govern Autonomous Recommender-Systems Research Labs},
  author={Beel, Joeran and Gipp, Bela and Vente, Tobias and Baumgart, Moritz and Meister, Philipp},
  journal={arXiv preprint arXiv:2510.18104},
  year={2025}
}

@inproceedings{beel2026autoreclabs,
  author    = {Beel, Joeran and Gipp, Bela and Vente, Tobias and
               Baumgart, Moritz and Meister, Philipp and Pourazari, Sinan},
  title     = {A {RecSys} Paper for \$20: Why We Must Build, Evaluate,
               and Govern Autonomous Recommender Systems Research Labs
               ({AutoRecLabs})},
  booktitle = {Methodology First -- Rethinking Research Assessment in RecSys
               (FRAME 2026)},
  year      = {2026},
  address   = {Minneapolis, Minnesota, USA},
  note      = {Accepted for publication}
}

@misc{Huang2023MLAgentBench,
  author    = {Huang, Qian and Vora, Jian and Liang, Percy and Leskovec, Jure},
  title     = {{MLAgentBench}: Evaluating Language Agents on Machine Learning Experimentation},
  year      = {2023},
  doi       = {10.48550/arXiv.2310.03302},
  publisher = {arXiv}
}

@misc{Chan2024MLEBench,
  author    = {Chan, Jun Shern and Chowdhury, Neil and Jaffe, Oliver and Aung, James and Sherburn, Dane and Mays, Evan and Starace, Giulio and Liu, Kevin and Maksin, Leon and Patwardhan, Tejal and Weng, Lilian and M{\k{a}}dry, Aleksander},
  title     = {{MLE-bench}: Evaluating Machine Learning Agents on Machine Learning Engineering},
  year      = {2024},
  doi       = {10.48550/arXiv.2410.07095},
  publisher = {arXiv}
}

@misc{Jiang2025AIDE,
  author    = {Jiang, Zhengyao and Schmidt, Dominik and Srikanth, Dhruv and Xu, Dixing and Kaplan, Ian and Jacenko, Deniss and Wu, Yuxiang},
  title     = {{AIDE}: AI-Driven Exploration in the Space of Code},
  year      = {2025},
  doi       = {10.48550/arXiv.2502.13138},
  publisher = {arXiv}
}

@inproceedings{Anand2020AutoSurprise,
  author    = {Anand, Rohan and Beel, Joeran},
  title     = {Auto-Surprise: An Automated Recommender-System ({AutoRecSys}) Library with Tree of Parzens Estimator ({TPE}) Optimization},
  booktitle = {Proceedings of the 14th ACM Conference on Recommender Systems},
  year      = {2020},
  pages     = {585--587},
  doi       = {10.1145/3383313.3411467}
}

@inproceedings{Vente2023LensKitAuto,
  author    = {Vente, Tobias and Ekstrand, Michael D. and Beel, Joeran},
  title     = {Introducing LensKit-Auto, an Experimental Automated Recommender System ({AutoRecSys}) Toolkit},
  booktitle = {Proceedings of the 17th ACM Conference on Recommender Systems},
  year      = {2023},
  pages     = {1212--1216},
  doi       = {10.1145/3604915.3610656}
}

\end{document}